\documentclass[10pt,twocolumn,letterpaper]{article}

\usepackage{cvpr}              

\usepackage[accsupp]{axessibility}
\usepackage{graphicx}
\usepackage{amsmath}
\usepackage{amssymb}
\usepackage{booktabs}
\usepackage{bm}
\usepackage{array}

\usepackage{multirow}
\usepackage{multicol}
\newcommand{\up}[1]{\tiny\textcolor{blue}{#1}}
\newcommand{\down}[1]{\tiny\textcolor{red}{#1}}
\usepackage[pagebackref,breaklinks,colorlinks]{hyperref}
\usepackage[capitalize]{cleveref}
\crefname{section}{Sec.}{Secs.}
\Crefname{section}{Section}{Sections}
\Crefname{table}{Table}{Tables}
\crefname{table}{Tab.}{Tabs.}

\def\cvprPaperID{5123} 
\def\confName{CVPR}
\def\confYear{2023}

\makeatletter
\def\thanks#1{\protected@xdef\@thanks{\@thanks
\protect\footnotetext{#1}}}
\makeatother

\begin{document}

\title{Partial Network Cloning}

\author{\bf Jingwen Ye \quad 
Songhua Liu \quad
Xinchao Wang$^{\dagger}$ \thanks{ $^{\dagger}$ Corresponding author.}\\
National University of Singapore\\
{\tt\small jingweny@nus.edu.sg, songhua.liu@u.nus.edu, xinchao@nus.edu.sg}\\
}
\maketitle

\begin{abstract}
In this paper, we study a novel task
that enables partial knowledge transfer 
from pre-trained models,
which we term as Partial Network Cloning~(PNC).
Unlike prior methods that
update  
all or at least part of the
parameters 
in the target network
throughout the knowledge transfer process, 
PNC conducts partial parametric ``cloning''
from a source network
and then injects the cloned 
module to the target, without modifying
its  parameters.
Thanks to the transferred module,
the target network is expected to gain
additional functionality, such as inference
on new classes;
whenever needed,
the cloned module can be readily removed
from the target,
with its original parameters
and competence
kept intact.
Specifically, we introduce 
an innovative
learning scheme that allows
us to identify simultaneously
the component to be cloned
from the source and the 
position to be inserted within
the target network,
so as to ensure the optimal performance.
Experimental results on several datasets
demonstrate that,
our method yields 
a significant improvement of $\>5\%$
in accuracy
and 50\%
in locality 
when compared 
with
{parameter-tuning based methods.}
Our code is available at \href{https://github.com/JngwenYe/PNCloning}{https://github.com/JngwenYe/PNCloning}.

\end{abstract}

\section{Introduction}
\label{sec:intro}

With the recent advances in deep learning, an 
increasingly number of pre-trained models have been
released online, demonstrating favourable performances on
various computer vision applications.
As such, many model-reuse approaches have been proposed
to take advantage of the pre-trained models.
In practical scenarios, users may request
to aggregate partial functionalities from multiple 
pre-trained networks, and customize
a target network whose competence
differs from any network in the model zoo.



\begin{figure}[t]
\centering
\includegraphics[scale = 0.55]{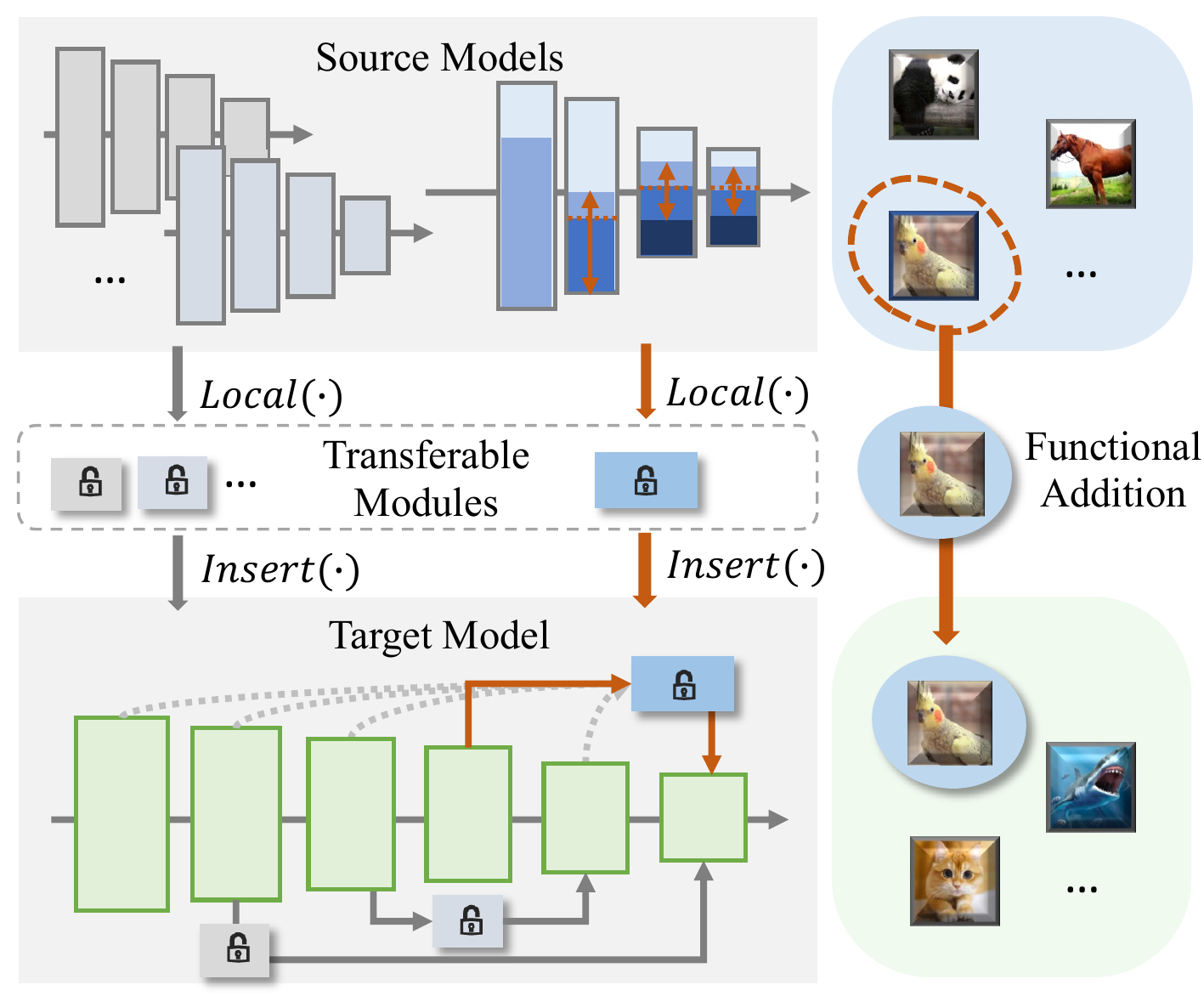}
\caption{Illustration of partial network cloning.
Given a set of pre-trained source models, 
we ``clone'' the transferable modules from the source, 
and insert them into the target model (left)
while preserving the functionality (right).
}
\label{fig:clone}
\end{figure}

A straightforward solution to the functionality dynamic changing is to re-train 
the target network using 
the original training dataset,
or to conduct finetuning together with 
{regularization strategies to alleviate catastrophic forgetting}~\cite{Li2016LearningWF,Chaudhry2020ContinualLI,Titsias2020FunctionalRF},
which is known as continual learning.
However, direct re-training is extremely inefficient,
let alone the fact that  original training dataset is often  unavailable.
Continual learning, on the other hand,
is prone to 
catastrophic forgetting especially
when the amount of data for finetuning is small,
which, unfortunately, often occurs in practice.
Moreover, both strategies 
inevitably overwrite the original parameters
of the target network, indicating that,
without explicitly storing original parameters of the target network,
there is no way to recover 
its original performance or competence when
this becomes necessary.

In this paper, we investigate
a novel task, termed as \textbf{P}artial \textbf{N}etwork \textbf{C}loning~(PNC),
to migrate knowledge from the source network,
in the form of a transferable module,
to the target one.
Unlike  prior methods that rely on updating 
parameters of the target network,
PNC attempts to \emph{clone}
partial parameters from the source network
and then directly inject the cloned module into the target,
as shown in Fig.~\ref{fig:clone}.
In other words, the cloned module
is transferred to the target
in a copy-and-paste manner.
Meanwhile, the original parameters of the
target network remain intact, indicating that
whenever necessary, the newly added 
module can be readily removed
to fully recover its original functionality.
Notably, the cloned module 
\emph{per se}
is a fraction of the source network, 
and therefore requirements 
no additional storage expect for the lightweight adapters.
Such flexibility to expand the 
network functionality
and to detach the cloned module
without altering the base of the target or 
{allocating extra storage},
in turn, greatly enhances the utility
of pre-trained model zoo
and largely enables plug-and-play
model reassembly.

Admittedly, the ambitious goal of PNC
comes with significant challenges,
mainly attributed to the 
black-box nature of the neural networks,
alongside our intention to preserve the 
performances on
both the previous and newly-added tasks of the target.
The first challenge concerns the localization
of the to-be-cloned module
within the source network, since
we seek discriminant representations
and good transferability 
to the downstream target task.
The second challenge, on the other hand,
lies in how to inject the cloned module
to ensure the performance.

To solve these challenges, 
we introduce an innovative strategy for PNC,
through learning the localization
and insertion in an intertwined manner
between the source and target network.
Specifically, 
to localize the transferable module
in the source network, we adopt a local-performance-based 
pruning scheme for parameter selection.
To adaptively insert the module into the target network, 
we utilize a positional search method 
in the aim to achieve the optimal performance, 
which, in turn, optimizes the localization operation.
The proposed PNC scheme
achieves performances
significantly superior to those of 
the continual learning setting ($5\%\sim 10\%$), 
while reducing  data dependency to $30\%$.

Our contributions are therefore summarized as follows.
\begin{itemize}
    \item We introduce a novel yet practical model re-use setup, termed as partial network cloning~(PNC). 
    In contrast to conventional settings
    the rely on updating all or part of
    the parameters in the target network, 
    PNC migrates parameters from the source  
    in a copy-and-paste manner to the target,
    while preserving original parameters of the target
    unchanged.
    
    
    
    
    \item We propose an effective scheme towards solving PNC, 
    which conducts learnable localization and insertion of
    the transferable module jointly between the source and target network.
    The two operations reinforce each other  
    and together ensure the performance of the target network.

    
    \item We conduct experiments on {four widely-used} datasets and 
    showcase that the proposed method consistently  achieves 
    results superior to the conventional knowledge-transfer settings,
    including continual learning and {model ensemble}.

\end{itemize}

\section{Related Work}
\subsection{Life-long Learning}
Life-long/online/incremental learning, which is capable of learning, retaining and transferring knowledge over a lifetime, has been a long-standing research area in many fields~\cite{Wu2018MemoryRG,Shmelkov2017IncrementalLO,ye2022learning,Ye2020DataFreeKA}. 
The key of continual learning is to solve catastrophic forgetting, and there are three main solutions, which are the regularization-based methods~\cite{Li2016LearningWF,Chaudhry2020ContinualLI,Titsias2020FunctionalRF,Huihui21AAAI}, the rehearsal-based methods~\cite{Choi2019AutoencoderBasedIC,Shin2017ContinualLW,Venkatesan2017ASF} and architecture-based methods~\cite{mallya2018piggyback,li2019learn,Yan_2021_CVPR,Kang2022ForgetfreeCL}. 

Among these three streams of methods, the most related one to PNC is the architecture-based pruning, which aims at minimizing the inter-task interference via newly designed architectural components. Li \textit{et al.}~\cite{li2019learn} 
propose to separate the explicit neural structure learning and the parameter estimation, and apply evolving neural structures to alleviate catastrophic forgetting. At each incremental step, DER~\cite{Yan_2021_CVPR} freezes the previously learned representation and augment it with additional feature dimensions from a new learnable feature extractor.
Singh \textit{et al.}~\cite{singh2020calibrating} choose to calibrate the activation maps produced by each network layer using spatial and channel-wise calibration modules and train only these calibration parameters for each new task.

The above incremental methods are fine-tuning all or part of the current network to solve functionality changes.
Differently, we propose a more practical life-long solution, which learns to transfer the functionality from pre-trained networks instead of learning from the new coming data.

\begin{figure*}[t]
\centering
\includegraphics[scale = 0.56]{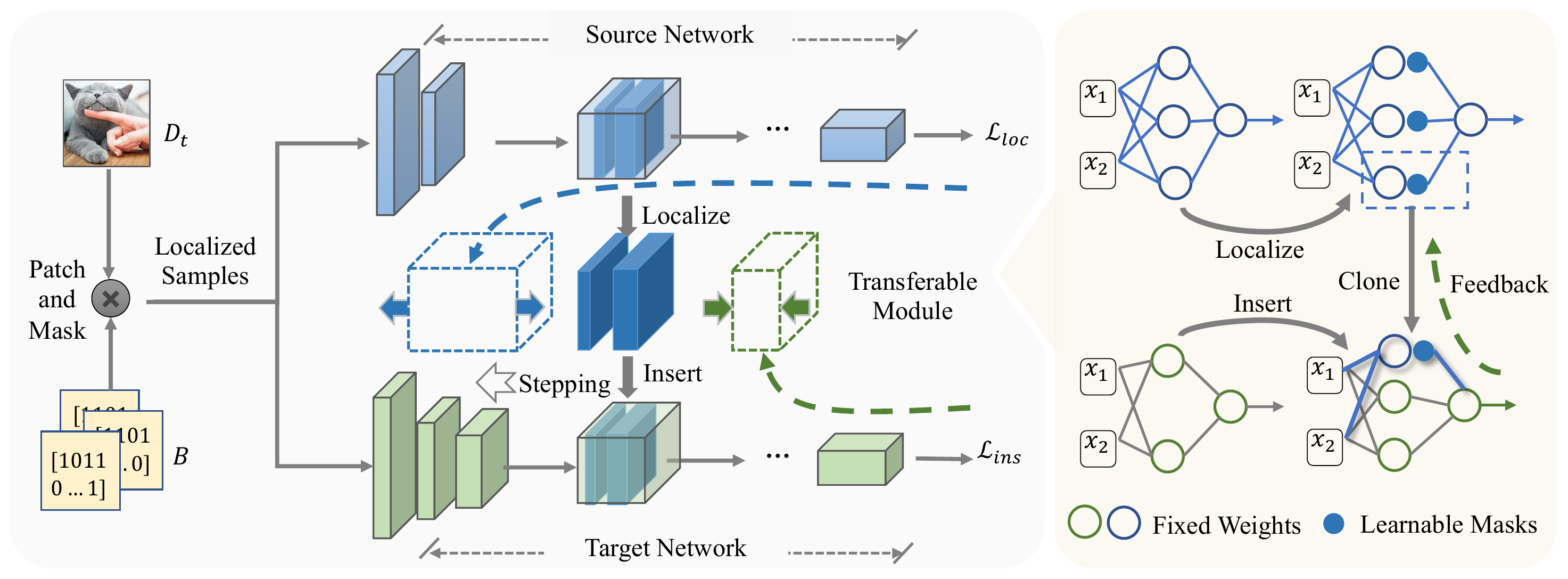}
\caption{The proposed partial network cloning framework.
The localized samples are fed into the source network for the original transferable module localization. To refine the transferable module, we learn how to locate and insert it, with the network weights fixed.}
\label{fig:framework}
\end{figure*}

\subsection{Network Editing}

Model editing is proposed to fix the bugs in networks, which aims to enable fast, data-efficient updates to a pre-trained base model’s behavior for only a small region of the domain, without damaging model performance on other inputs of interest~\cite{sinitsin2020editable, mitchell2022memory, sotoudeh2021provable}.

A popular approach to model editing is to establish learnable
model editors, which are trained to predict updates to the
weights of the base model to produce the desired change in behavior~\cite{sinitsin2020editable}. MEND~\cite{mitchell2021fast} utilizes a collection of small auxiliary editing networks as a model editor.
Eric \textit{et al.}~\cite{mitchell2022memory} propose to store edits in an explicit memory and learn to reason over them to modulate the base model’s predictions as needed. 
Provable point repair algorithm \cite{sotoudeh2021provable} finds a provably minimal repair satisfying the safety specification over a finite set of points. 
Cao \textit{et al.}~\cite{de2021editing} propose to train a hyper-network with constrained optimization to modify without affecting the rest of the knowledge, which is then used to predict the weight update at test time. 

Different from network edition that directly modifies a certain of weights to fix several bugs, our work do the functionality-wise modification by directly inserting the transferable modules.

\subsection{Model Reuse}
With a bulk of pre-trained models online, model reuse becomes a hot topic, which attempts to construct the model by utilizing existing available models, rather than building a model from scratch. 
Model reuse is applied for the purpose of reducing the time complexity, data dependency or and expertise requirement, which is studied by knowledge transfer~\cite{Han2020NeuralCM,FangPrunging2023CVPR,gao2017knowledge,Yuan2020CKDCK,Ye2021SafeDB,Ren_2021_CVPR} and model ensemble~\cite{peng2022model,shen2019meal,cao2020ensemble,walawalkar2020online}.

Knowledge transfer~\cite{Ren_2022_CVPR,liu2022dataset,yang2022KF,YangKD2023CVPR,yang2020CVPR,yang2020NeurIPS,LiuStyle2022ECCV} utilizes the pre-trained models by transferring knowledge from these networks to improve the current network, which has promoted the performance of domain adaptation~\cite{jing2020adaptively}, multi-task learning~\cite{wang2021coarse},  Few-Shot Learning~\cite{Li_2019_CVPR} and so on~\cite{jing2022learning}.
For example, KTN~\cite{Peng_2019_ICCV} is proposed to jointly incorporate visual feature learning, 
knowledge inferring and classifier learning into one unified framework for their optimal compatibility.
To enable transferring knowledge from multiple models, 
Liu \textit{et al.} ~\cite{liu2020adaptive} 
propose an adaptive multi-teacher multi-level knowledge distillation learning framework which associates each teacher with a latent representation to adaptively learn the importance weights.

Model ensemble~\cite{jing2021amalgamating,jing2021meta,yang2022deep} integrates multiple pre-trained models to obtain a low-variance and generalizable model. Peng \textit{et al.}~\cite{peng2022model} apply sample-specific ensemble of source models by adjusting the contribution of each source model for each target sample. MEAL~\cite{shen2019meal} proposes an adversarial-based learning strategy in block-wise training to distill diverse knowledge from different trained models.

The above model reuse methods transfer knowledge from networks to networks, with the base functionality unchanged. 
We make the first step work to directly transfer part of the knowledge into a transferable module by cloning part of the parameters from the source network, which enables network functionality addition.

\section{Proposed Method}

The goal of the proposed partial network cloning framework is to clone part of the source networks to the target network so as to enable the corresponding functionality transfer in the target network.

The illustration of the proposed PNC framework is shown in Fig.~\ref{fig:framework}, where we extract a transferable module that could be directly inserted into the target network.

\subsection{Preliminaries}

Given a total number of $P$ pre-trained source models $\mathcal{M}_s = \{\mathcal{M}_{s}^0, \mathcal{M}_{s}^1,...,\mathcal{M}_{s}^{P-1}\}$, 
each $\mathcal{M}_{s}^{\rho}$ ($0\le \rho< P$) serves for cloning the functionality $t_s^{\rho}$, where $t_s^{\rho}$ is a subset of the whole functionality set of $\mathcal{M}_{s}^{\rho}$ and the to-be-cloned target set is denoted as $T_s=\{t_s^{0}, t_s^{1}, ..., t_s^{P-1}\}$.
The partial network cloning is applied on the target model $\mathcal{M}_t$ for new functionalities addition, which is the pre-trained model on the original set $T_t$ ($T_t\cap T_s = \emptyset$). 

Partial network cloning aims at expending the functionality set of target network on the new $T_s$ by directly cloning.
In the proposed framework, it is achieved by firstly extracting part of $\mathcal{M}_s$ to form a transferable module $\mathcal{M}_f$, and then inserting it into target model $\mathcal{M}_t$ to build a after-cloned target network $\mathcal{M}_c$. 
The whole process won't change any weights of the source and target models, and also each transferable module is directly extracted from the source model free of any tuning on its weights. 
Thus, the process can be formulated as:
\begin{equation}
\begin{split}
    \mathcal{M}_c \leftarrow Clone(\mathcal{M}_t,M,\mathcal{M}_s,R ),
    \label{eq:clone}
\end{split}
\end{equation}
which is directly controlled by $M$ and $R$, where $M$ is a set of selection functions  for deciding how to extract the explicit transferable module on source networks $\mathcal{M}_s$, and $R$ is the position parameters for deciding where to insert the transferable modules to the target network. 
Thus, partial network cloning $Clone(\cdot)$ consists of two steps:
\begin{equation}
    \begin{split}
            \mathcal{M}_f^{\rho} &\leftarrow Local(\mathcal{M}_s^{\rho}, M^{\rho}),\\
    \mathcal{M}_c &\leftarrow Insert_{\rho=0}^P(\mathcal{M}_t,\mathcal{M}_f^{\rho},R^{\rho}),    \end{split}
\end{equation}
where both $M$ and $R$ are learnable and optimized jointly. Once $M$ and $R$ are learned, $\mathcal{M}_c$ can be determined with some lightweight adapters. 

Notably, 
we assume that only the samples related to the to-be-cloned task set $T_s$ are available in the whole process, keeping the same setting of continual learning.
And to be practically feasible, partial network cloning must meet three natural requirements:
\begin{itemize}
    \item \textbf{Transferability:} The extracted transferable module should contain the explicit knowledge of the to-be-cloned task $T_s$, which could be transferred effectively to the downstream networks;
    \item \textbf{Locality:} The influence on the cloned model $\mathcal{M}_c$ out of the target data $D_t$ should be minimized;
    \item \textbf{Efficiency:} Functional cloning should be efficient in terms of runtime and memory;
    \item \textbf{Sustainability:} The process of cloning wouldn't do harm to the model zoo, meaning that no modification the pre-trained models are allowed and the cloned model could be fully recovered.
\end{itemize}

In what follows, we consider the partial network cloning from one pre-trained network to another, which could certainly be extended to the multi-source cloning cases, thus we omit $\rho$ in the rest of the paper.



\if
In the framework of network partial cloning, there are three main steps:
\begin{itemize}
    \item \textbf{Step 1}: We need to localize the part of the network that are mostly related to the target function or task, which is achieved by locally disturbing the input.
    \item \textbf{Step 2}: Insert and optimize.
    \item \textbf{Step 3}: Adversarial adjustment with step 1 and step 2.
\end{itemize}
\fi

\subsection{Localize with pruning}
Localizing the transferable module from the source network is actually to learn the selection function $M$. 

In order to get an initial transferable module $\mathcal{M}_f$, we locate the explicit part in the source network $\mathcal{M}_s$ that contributes most to the final prediction. Thus, the selection function $M$ is optimized by the transferable module's performance locally on the to-be-cloned task $T_s$.

Here, we choose the selection function as a kind of mask-based pruning method mainly for two purposes: the first one is that it applies the binary masks on the filters for pruning without modifying the weights of $\mathcal{M}_s$, thus, ensuring sustainability; 
the other is for transferability that pruning would be better described as `selective knowledge damage'~\cite{Hooker2020WhatDC}, which helps for partial knowledge extraction.

Note that unlike the previous pruning method with the objective function to minimize the error function on the whole task set of $\mathcal{M}_s$,
here, the objective function is designed to minimize the \textit{locality performance} on the to-be-cloned task set $T_s$.
Specifically, for the source network $\mathcal{M}_s$ with $L$ layers $W_s = \{w_s^0, w_s^1, ..., w_s^{L-1}\}$, the localization can be denoted as:
\begin{equation}
\begin{split}
    \mathcal{M}_f =& M\cdot\mathcal{M}_s \Leftrightarrow \{m^l\cdot w_s^l|0\le l<L\},\\
    M =& \arg\max_M \ Sim(\mathcal{M}_f,\mathcal{M}_s|D_t)\\
    &- Sim(\mathcal{M}_f, \mathcal{M}_s|\overline{D}_{t}),
\label{eq:local}
\end{split}
\end{equation}
where $M=\{m^0, m^1, ..., m^{L-1}\}$ is a set of learnable masking parameters, which are also the selection function as mentioned in Eq.~\ref{eq:clone}. $Sim(\cdot|\cdot)$ represents the conditional similarity among networks, $\overline{D}_{t}$ is the rest data set of the source network. The localization to extract the explicit part on the target $D_t$ is learned by maximizing the similarity between $\mathcal{M}_s$ and $\mathcal{M}_f$ on $D_t$ while minimizing it on $\overline{D}_{t}$.

Considering the black-box nature of deep networks that all the knowledge (both from $D_t$ and $\overline{D}_t$) is deeply and jointly embedded in $\mathcal{M}_s$, it is non-trivial to calculate the similarity on the ${D_t}$-neighbor source network $\mathcal{M}_s|_{D_t}$. Motivated by LIME~\cite{Ribeiro2016WhySI} 
that utilizes interpretable
representations locally faithful to the classifier,
we train a model set containing $N$ small local models $\mathcal{G}=\{g_i\}^{(N)}$ to model the source $\mathcal{M}_s$ in the $D_t$ neighborhood, and then use the local model set as the surrogate: $\mathcal{G}\approx \mathcal{M}_s|_{D_t}$.
To obtain $\mathcal{G}$, for each $x_i \in D_t$, we get its augmented neighborhood by separating it into patches (i.e. $8\times 8$) and applying the patch-wise perturbations with a set of binary masks $B$. 
Thus, $\mathcal{G}$ is obtained by:
\begin{equation}
\min_{g_i} \frac{1}{|B|}\sum_{b\in B}\Pi_b\cdot\big\|\mathcal{M}_s(b\cdot x_i)-g_i(b))\big\|^2+\Omega(g_i),    
\end{equation}
where $\Pi_{b}$ is the weight measuring sample locality according to ${x}_i$,
$\Omega(g_i)$ is the complexity of $g_i$ and $|B|$ donates the total number of masks. 
$\mathcal{G}$ is optimized by the least square method and more details are given in the supplementary.
For each $x_i$, we calculate a corresponding $g_i$. And actually, we set $N<|D_t|$ (about $30\%$), which is clarified in the experiments.

The new $\mathcal{G}$, 
{calculated from the original source network $\mathcal{M}_s$ in the $D_t$ neighborhood}, models the locality of the target task $T_s$ on $\mathcal{M}_s$. Note that $\mathcal{G}$ can be calculated in advance for each pre-trained model, as it could also be a useful tool for the model distance measurement and others~\cite{Ilyas2022DatamodelsPP}.
In this paper, $\mathcal{G}$ perfectly matches our demand for the transferable module localization.
So the localization process in Eq.~\ref{eq:local} could be optimized as:
\begin{equation}
\begin{split}
 \min_{M}\sum_{g_i\in \mathcal{G}}\sum_{b\in B} \Big\|& f_t[\mathcal{M}_s\big(M\cdot W_s ; b\cdot x\big)] - f_t[g_i(b)]\Big\|^2\\
&s.t.\quad |m^l|\le c^l
\label{eq:locloss}
\end{split}
\end{equation}
where $f_t$ is for selecting the $T_s$ related output and $c^l$ is the parameter controlling the number of non-zero values of $M$ ($c^l<|W^l|$). 
And for inference, the learned soft masks $M$ are binarized by selecting $c^l$ filters with the top-$c^l$ masking values in each layer.

\subsection{Insert with adaptation }
After the transferable module $\mathcal{M}_f$ being located at the source network, it could be directly extracted from $\mathcal{M}_s$ with $M$, without any modifications on its weights. 
Then the following step is to decide where to insert $\mathcal{M}_f$ into $\mathcal{M}_t$, as to get best insertion performance.

The insertion is controlled by the position parameter $R$ mentioned in Eq.~\ref{eq:local}.
Following most of the model reuse settings that keep the first few layers of the pre-trained model as a general feature extractor, the learning-to-insert process with $R$ is simplified as finding the best position ($R$-th layer to insert $\mathcal{M}_f$).
The insertion could be denoted as:
\begin{equation}
\begin{split}
 \mathcal{M}_c^R =& \mathcal{M}_t\big(W_t^{[0:R]}\big)\circ \big\{\mathcal{M}_t\big(W_t^{[R:L]}\big),\mathcal{M}_f\big\},\\
R^* =&\arg\max_R \ Sim(\mathcal{M}_f,\mathcal{M}_c^R|D_t)\\
&+ Sim(\mathcal{M}_t,\mathcal{M}_c^R|D_o),
\label{eq:insert}
\end{split}
\end{equation}
where $D_o$ is the original set for pre-training the target network, and $D_o\cup D_t = \emptyset$. 
The cloned $\mathcal{M}_c$ is obtained by the parallel connection of the transferable module into the target network $\mathcal{M}_t$.
Thus the insertion learned by Eq.~\ref{eq:insert} is to find the best insertion position by maximizing the similarity between $\mathcal{M}_f$ and $\mathcal{M}_c$ on $D_t$ (for the best insertion performance on $D_t$) and the similarity between $\mathcal{M}_t$ and $\mathcal{M}_c$ on $D_o$ (for the least accuracy drop on the previously learned $D_o$).

In order to learn the best position R, we need maximize the network similarities $Sim(|)$. Different from the solution used to optimize the objective function while localizing, insertion focuses on the prediction accuracies on the original and the to-be-cloned task set. So we use the network outputs to calculate $Sim(\cdot)$, which is the KL-divergence loss $\mathcal{L}_{kd}$. we write:
\begin{equation}
\begin{split}
       \min_{\mathcal{F}_c, \mathcal{A}}&\mathcal{L}_{kd}\circ f_t\big[\mathcal{F}_c(\mathcal{A};\mathcal{M}_c^R(B\cdot x)), \mathcal{G}(B)\big]\\
       +&\mathcal{L}_{kd}\circ \overline{f}_t\big[\mathcal{F}_c(\mathcal{A};\mathcal{M}_c^R(B\cdot x)), \mathcal{M}_t(B\cdot x)\big],\\
       &s.t.\quad R \in \{0,1,...,L-1\}
\label{eq:rtrain}
\end{split}
\end{equation}
where $f_t$ is for selecting the $T_s$ related output while $\overline{f}_t$ is for selecting the rest. 
$\mathcal{F}_c$ is the extended fully connection layers from the original FC layers of $\mathcal{M}_t$. And we add an extra adapter module $\mathcal{A}$ to do the feature alignment for the transferable module, which further enables cloning between heterogeneous models.
The adapter is consisted of one $1\times 1$ conv layer following with ReLu, which, comparing with $\mathcal{M}_s$ and $\mathcal{M}_t$, is much smaller in scale.
$\mathcal{G}$ and $B$ are defined in Eq.~\ref{eq:locloss}.

While training, $R$ is firstly set to be $L-1$ and then moving layer by layer to $R=0$.
In each moving step, we fine-tune the adapter $\mathcal{A}$ and the corresponding fully connected layers $\mathcal{F}_c$. It is a light searching process, since only a few of weights ($\mathcal{A}$ and $\mathcal{F}_c$) need to be fine-tuned for only a couple epochs (5$\sim$ 20). Extra details for heterogeneous model pair are in the supplementary.
Please note that although applying partial network cloning from the source to the target needs two steps ($Clone(\cdot)$ and $Insert(\cdot)$), the learning process is not separable and are interacted on each other.
As a result, the whole process can be jointly formulated as:
\begin{equation}
\begin{split}
 \!\min_{M,\mathcal{F}_c,\mathcal{A}}\mathcal{L}_{loc}\Big(\mathcal{M}_s^{[0:R]}&, M\!\cdot\! W_s^{[R:L]}, \mathcal{G}\Big)\!+\! \mathcal{L}_{ins} \Big( \mathcal{M}_t^{[0:R]}, \\
 \mathcal{A}\circ(M&\!\cdot\! W_s^{[R:L]}), \mathcal{M}_t^{[R:L]}, \mathcal{F}_c, \mathcal{G}\Big),\\
where &\quad R: (L-1) \rightarrow 0 
\end{split}
\label{eq:lossall}
\end{equation}
where $\mathcal{L}_{loc}(\cdot)$ is the objective function in Eq.~\ref{eq:locloss} and $\mathcal{L}_{ins}(\cdot)$ is the objective function in Eq.~\ref{eq:rtrain}. And in this objective function, $\mathcal{M}_s$ and $\mathcal{M}_t$ are using the same $R$ for simplification, while in practice a certain ratio exists for the heterogeneous model pair. 

Once the above training process is completed, we could roughly estimate the performance by the loss convergence value, which follows the previous work~\cite{Ye_Amalgamating_2019}.
Finally the layer with least convergence value is marked as the final $R$. The insertion is completed by this determined $R$ and the corresponding $\mathcal{A}$ and $\mathcal{F}_c$.

\subsection{Cloning in various usages}
The proposed partial network cloning by directly inserting a fraction of the source network enables flexible reuse of the pre-trained models in various practical scenarios.\\
\textit{\textbf{Scenario I:} Partial network cloning is a better form for information transmission.} When there is a request for transferring the networks, it is better to transfer the cloned network obtained by PNC as to reduce latency and transmission loss. 

In the transmission process, we only need to transfer the set $\{M,R,\mathcal{A},\mathcal{F}_c\}$, which together with the public model zoo, could be recovered by the receiver. 
$\{M,R,\mathcal{A},\mathcal{F}_c\}$ is extremely small in scale comparing with a complete network, thus could reduce the transmission latency.
And if there is still some transmission loss on $\mathcal{A}$ and $\mathcal{F}_c$, 
it could be easily revised by the receiver by fine-tuning on $D_t$. 
As a result, PNC provides a new form of networks for high-efficiency transmission.\\
\textit{\textbf{Scenario II:} Partial network cloning enables model zoo online usage.} In some resource limited situation, the users could flexibly utilize model zoo online without downloading it on local.

Note that the cloned model is determined by $Clone(\mathcal{M}_t,M,\mathcal{M}_s,R )$, $\mathcal{M}_t$ and $\mathcal{M}_s$ are fixed and unchanged in the whole process. There is not any modifications on the pre-trained models ($\mathcal{M}_t$ and $\mathcal{M}_s$) nor introducing any new models. PNC enables any functional combinations in the model zoo,
which also helps maintain a good ecological environment for the model zoo, 
since PNC with $M$ and $R$ is a simple masking and positioning operation, which is easy of revocation. Thus, the proposed PNC supports to establish a sustainable model zoo online inference platform.

\section{Experiments}

We provide the experimental results on four publicly available benchmark datasets, and evaluate the cloning performance in the commonly used metrics as well as the locality metrics.
And we compare the proposed method with the most related field -- continual learning, to show concrete difference between these two streams of researches.  
More details and experimental results including partially cloning from multiple source networks, can be found in the supplementary.

\subsection{Experimental settings}
\textbf{Datasets.}
Following the setting of previous continual methods, we report experiments on MNIST, CIFAR-10, CIFAR-100 and TinyImageNet datasets.
For MNIST, CIFAR-10 and CIFAR-100 datasets, we are using input size of $32\times 32$.
For TinyImageNet dataset, we are using input size of $256\times 256$.
In the normal network partial cloning setting, 
the first 50\% of classes are selected to pre-train the target network $\mathcal{M}_t$, and the last 50\% of classes classes are selected to pre-train the source network $\mathcal{M}_s$.

In the partial network cloning process, $30\%$ of the training data are used for each sub dataset, which reduces the data dependency to 30\%. And for training the local model set $\mathcal{G}$, we set $|B|=100$ and segment the input into $4\times 4$ patches for the MNIST, CIFAR-10 and CIFAR-100 datasets, set $|B|=1000$ and segment the input into $8\times 8$ patches for the Tiny-ImageNet dataset.

\textbf{Training Details.}
We used PyTorch framework for the implementation.
We apply the experiments on the several network backbones, including plain CNN, LeNet, ResNet, MobileNetV2 and ShuffleNetV2.
In the pre-training process, we employ a standard data augmentation strategy: random crop, horizontal flip, and rotation.
In the process of partial cloning, 10 epochs fine-tuning are operated for each step on MNIST and CIFAR-10 datasets, 20 epochs for CIFAR-100 and Tiny-ImageNet datasets.

For simplifying and accelerating the searching process in Eq.~\ref{eq:lossall}, we split LeNet into 3 blocks, the ResNet-based network into 5 blocks, MobileNetV2 into 8 blocks and ShuffleNetV2 into 5 blocks (excluding the final FC layers). Thus the block-wise adjustment for $R$ is applied for acceleration. 

\textbf{Evaluation Metrics.}
For the cloning performance evaluation, we evaluate the task performance by average accuracy:`Ori. Acc' (accuracy on the original set), `Tar. Acc' (accuracy on the to-be-cloned set) and `Avg. Acc' (accuracy on the original and to-be-cloned set), which is evaluated on the after-cloning target network $\mathcal{M}_c$. 

For evaluating the transferable module quality evaluation on local-functional representative ability, we use the conditional similarity $Sim(|)$ with $\mathcal{G}$~\cite{Jia2022AZO}, which can be calculated as:
\begin{equation}
    Sim(\mathcal{M}_a | D_a, \mathcal{M}_b | D_b) = Sim_{cos}\theta(\mathcal{G}_a,\mathcal{G}_b)
\end{equation}
where $Sim_{cos}(\cdot)$ is the cosine similarity, $\mathcal{G}_a$ and $\mathcal{G}_b$ are the corresponding local model sets of $\mathcal{M}_a(D_a)$ and $\mathcal{M}_b(D_b)$.

For evaluating the transferable module quality on transferability to other networks other than the target network, it is in the supplementary. 

\subsection{Experimental Results}
\begin{table*}[!htb]
\centering
\scriptsize
\caption{Overall performance on partial network cloning on MNIST, CIFAR10, CIFAR100 and Tiny-ImageNet datasets. We report the accuracies to evaluate the performance, including the comparison with the other functional addition methods and the ablation study. We choose \textit{`Continual+KD' as baseline} and mark the accuracy promotion in blue, accuracy drop in red.}
\label{tab:overall}
\begin{tabular}{p{19mm}|p{8mm}<{\centering}p{8mm}<{\centering}p{8mm}<{\centering}|p{8mm}<{\centering}p{8mm}<{\centering}p{8mm}<{\centering}||p{8mm}<{\centering}p{8mm}<{\centering}p{8mm}<{\centering}|p{8mm}<{\centering}p{8mm}<{\centering}p{8mm}<{\centering}}
\toprule
& \multicolumn{6}{c||}{\textbf{Acc on MNIST (LeNet5, \#3 Steps)}}& \multicolumn{6}{c}{\textbf{Acc on CIFAR-10 (ResNet-18, \#5 Steps)}} \\
\textbf{Method}  &\textbf{Ori.-S}&  \textbf{Tar.-S}&\textbf{Avg.-S}& \textbf{Ori.-M}&  \textbf{Tar.-M}&\textbf{Avg.-M} &\textbf{Ori.-S}&  \textbf{Tar.-S}&\textbf{Avg.-S}& \textbf{Ori.-M}&  \textbf{Tar.-M}&\textbf{Avg.-M}\\\midrule
Pre-trained & 99.7 & 99.5 &99.7&99.7&99.5&99.6 &95.9&97.2&96.1 & 95.9 & 97.6& 96.5\\
Joint+Full Set & 99.8  &  98.3 & 99.6 &99.7 & 99.3 & 99.5&95.2 & 96.8& 95.5&94.4&95.1& 94.7\\
Continual&83.4\down{-10.1}&100.0\up{+17.3}& 86.2\down{-5.5} &65.1\down{-27.9}&98.8\up{+16.8}& 77.7\down{-11.2}& 67.7\up{+2.8}&97.2\up{+2.6}& 75.3\down{-14.8} & 92.8\up{+18.7}&78.2\up{+16.6}&87.3\down{-2.1} \\
Direct Ensemble &94.6\up{+1.1} & 56.1\down{-26.4} &88.2\down{-3.5} &94.6\up{+1.6}& 81.9\down{-0.1} & 89.8\up{+0.9} &90.5\up{+25.6}& 39.3\down{-55.3}& 82.0\up{+12.1} &90.5\up{+16.4}& 43.8\down{-17.8}&73.0\up{+3.6}\\
\textit{Continual+KD} & \textit{93.5} &\textit{82.7}& \textit{91.7}& \textit{93.0}&\textit{82.0}& \textit{88.9}& \textit{64.9} &\textit{94.6} & \textit{69.9} &\textit{74.1}&\textit{61.6}& \textit{69.4}\\
\midrule
PNC-F (w/o Local) & 87.7\down{-5.8} & 100.0+\up{17.3} &90.0\down{-1.7} & 90.9\down{-2.1} &98.2\up{+16.2} & 93.6\up{+4.7}& 88.6\up{+23.7} & \textbf{97.3}\up{+2.7} & 90.1\up{+20.2} &85.5\up{+11.4} & 95.8\up{+34.2} & 89.4\up{+20.0}\\
PNC-F (w/o Insert)   & 86.9\down{-6.6} &  100.0+\up{17.3} & 89.1\down{-2.6}& 90.4\down{-2.6} &97.7\up{+15.7} &93.1\up{+4.2} & 86.1\up{+21.2}& 96.8\up{+2.2} &87.9\up{+18.0} & 86.0\up{+11.9} & \textbf{96.2}\up{+34.6} & 89.8\up{+30.4} \\
PNC-F (full)   & 88.5\down{-5.0}&  \textbf{99.7}\up{+17.0}& 90.4\down{-2.6} & 91.1\down{-1.9} &98.8\up{+16.8} & 94.0\up{+5.1}& 83.0\up{+18.1} & 96.5\up{+1.9}&85.3\up{+15.4} & 85.4\up{+11.3}&95.5\up{+33.9}& 89.2\up{+19.8}\\
\midrule
PNC (w/o Local) & 93.6\up{+0.1} & 96.2\up{+13.5} &94.0\up{+2.3} &92.9\down{-0.1}&94.0\up{+12.0}& 93.3\up{+4.4}& 90.5\up{+25.6} & 93.9\down{-0.7} & 91.7\up{+21.8}&87.1\up{+13.0}& 94.6\up{+33.1}&89.9\up{+29.8}\\
PNC (w/o Insert) & 92.8\down{-0.7} & 99.5\up{+16.8} & 93.9\up{+2.2}&91.9\down{-1.1} & 97.3\up{+15.3}& 93.9\up{+5.0}& 89.5\up{+24.6}&94.4\down{-0.2}& 90.3\up{+20.4}& 89.2\up{+15.1}&94.7\up{+33.2}& 91.3\up{+21.9}\\
\textbf{PNC (Ours, full)} & \textbf{96.4}\up{+2.9} & \textbf{99.7}\up{+17.0} &\textbf{97.0}\up{+5.3} & \textbf{96.2}\up{+3.2} &\textbf{97.8}\up{15.8} & \textbf{96.8}\up{+7.9}& \textbf{94.9}\up{+30.0} & 95.5\up{+0.9}&\textbf{95.0}\up{+25.1} & \textbf{93.7}\up{+19.6}& 94.5\up{+32.9}& \textbf{94.0}\up{+24.6} \\\midrule\midrule
& \multicolumn{6}{c||}{\textbf{Acc on CIFAR-100 (ResNet-50, \#5 Steps)}}& \multicolumn{6}{c}{\textbf{Acc on Tiny-ImageNet ( ResNet-18, \#5 Steps)}} \\
\textbf{Method}  &\textbf{Ori.-S}&  \textbf{Tar.-S}&\textbf{Avg.-S}& \textbf{Ori.-M}&  \textbf{Tar.-M}&\textbf{Avg.-M} &\textbf{Ori.-S}&  \textbf{Tar.-S}&\textbf{Avg.-S}& \textbf{Ori.-M}&  \textbf{Tar.-M}&\textbf{Avg.-M}\\\midrule
Pre-trained &80.0&	80.3& 80.1&80.0 &77.2 & 79.0 & 71.3 & 67.6  & 70.7&  71.3 & 68.9 & 70.4\\
Joint+Full Set  & 78.0& 74.9&77.5 &	76.3&	77.9 &76.9 &  63.1 & 60.8 &62.7 &63.7&61.6 & 62.9 \\
Direct Ensemble & 59.3\down{-6.2}  &46.4\down{-26.3} &57.2\down{-9.6} & 56.0\down{-18.4} & 46.4\down{-26.6} &52.4\down{-21.5} &58.0\up{+0.8} & 35.9\down{-20.5} & 54.3\down{-2.8}& 50.6\down{-9.3} & 30.2\down{-27.9} &43.0\down{-16.3}\\
Continual & 52.3\down{-13.2} & \textbf{79.4}\up{+6.7} & 56.8\down{-9.9} & 58.8\down{-15.6} & \textbf{78.0}\up{+5.0} &66.0\down{-7.9} &54.6\down{-2.6} & \textbf{70.1}\up{+13.7} & 57.2\up{+0.1} & 55.9\down{-4.0} & \textbf{64.9}\up{+6.8}& 59.3\up{+0.1} \\
\textit{Continual + KD} & \textit{65.5} & \textit{72.7} &66.7 & \textit{74.4} & \textit{73.0} &73.9 &\textit{57.2} &\textit{56.4} & 57.1& \textit{59.9} & \textit{58.1} & 59.2\\\midrule
PNC (w/o Local) &72.2\up{+6.7} & 70.4\down{-2.3} &71.9\up{+5.2} &75.7\up{+1.3} & 68.3\down{-4.7} &72.9\down{-1.0} & \textbf{65.6}\up{+8.4} &52.5\down{-3.9} & \textbf{63.4}\up{+6.4} &56.4\down{-3.5} & 55.9\down{-2.2} & 56.2\down{-3.0}\\
PNC (w/o Insert)  &63.2\down{-2.3} & 76.1\up{+3.4} &65.4\down{-1.3} & 66.1\down{-8.3} &76.0\up{+3.0} &69.8\down{-4.1} & 60.7\up{+3.5} & 63.5\up{+7.1} &61.2\up{+4.1} &58.8\down{-1.1} & 60.9\up{+2.8} & 59.6\up{+0.4} \\
\textbf{PNC (Ours, full)} &\textbf{76.7}\up{+11.2} & 74.9\up{+2.2}  & \textbf{76.4}\up{+9.7}& \textbf{76.9}\up{+2.5}  &76.5\up{+3.5} & \textbf{76.8}\up{+2.9}& 63.2\up{+6.0} & 60.7\up{+4.3} &62.8\up{+5.7} & \textbf{63.5}\up{+3.6} &60.4\up{+2.3} & \textbf{62.3}\up{+3.1} \\
\bottomrule
\end{tabular}
\end{table*}
\subsubsection{Overall Performance}
Table~\ref{tab:overall} shows overall performance of partial network cloning on MNIST, CIFAR-10, CIFAR-100 and Tiny-ImageNet datasets, where the target network and the source network are set to be the same architecture and the number of search steps $R$ is also listed.
We compare the proposed partial network cloning (`PNC') with the baseline `Pre-trained' original networks ( Acc on $\mathcal{M}_s$ and $\mathcal{M}_t$), 
the ensemble network of the source and the target (`Direct Ensemble'),
the networks scratch trained on the set including the original and the target (`Joint + Full set'), 
the continual-learned network with some regularization item (`Continual') and the continual-learned network with KD loss from the source network (`Continual+KD'). Specially, we compare the proposed `PNC' with `PNC-F', where `PNC-F' is the afterward all-parameter-tuned version of `PNC' on the to-be-cloned dataset.
And we also give the comparisons on the small-scale functionality addition (`-S', 20\% of the source functionalities are transferred), and middle-scale functionality addition (`-S', 60\% of the source functionalities are transferred).

From Table~\ref{tab:overall}, several observations are obtained.
Firstly, {the proposed PNC is capable of dealing with various datasets and network architectures} and its effectiveness has been proved on four datasets and on different network architectures.
Secondly, the full setting PNC gives the best solution to the new functionality addition task, our full setting (`PNC(full)') outperforms almost all of the other methods.
Thirdly, PNC shows better performance when cloning smaller functionality (`Avg.-S' vs `Avg.-M'), and it is practical in use when the most related network is chosen as the target and minor functionality is added with the proposed PNC. 
Finally, fully fine-tuning all the parameters of $\mathcal{M}_c$ after PNC doesn't bring any benefits (`PNC' vs `PNC-F'), since fine-tuning with the to-be-cloned dataset would bring bias on the new functionality.

\begin{figure}[t]
\centering
\includegraphics[scale = 0.44]{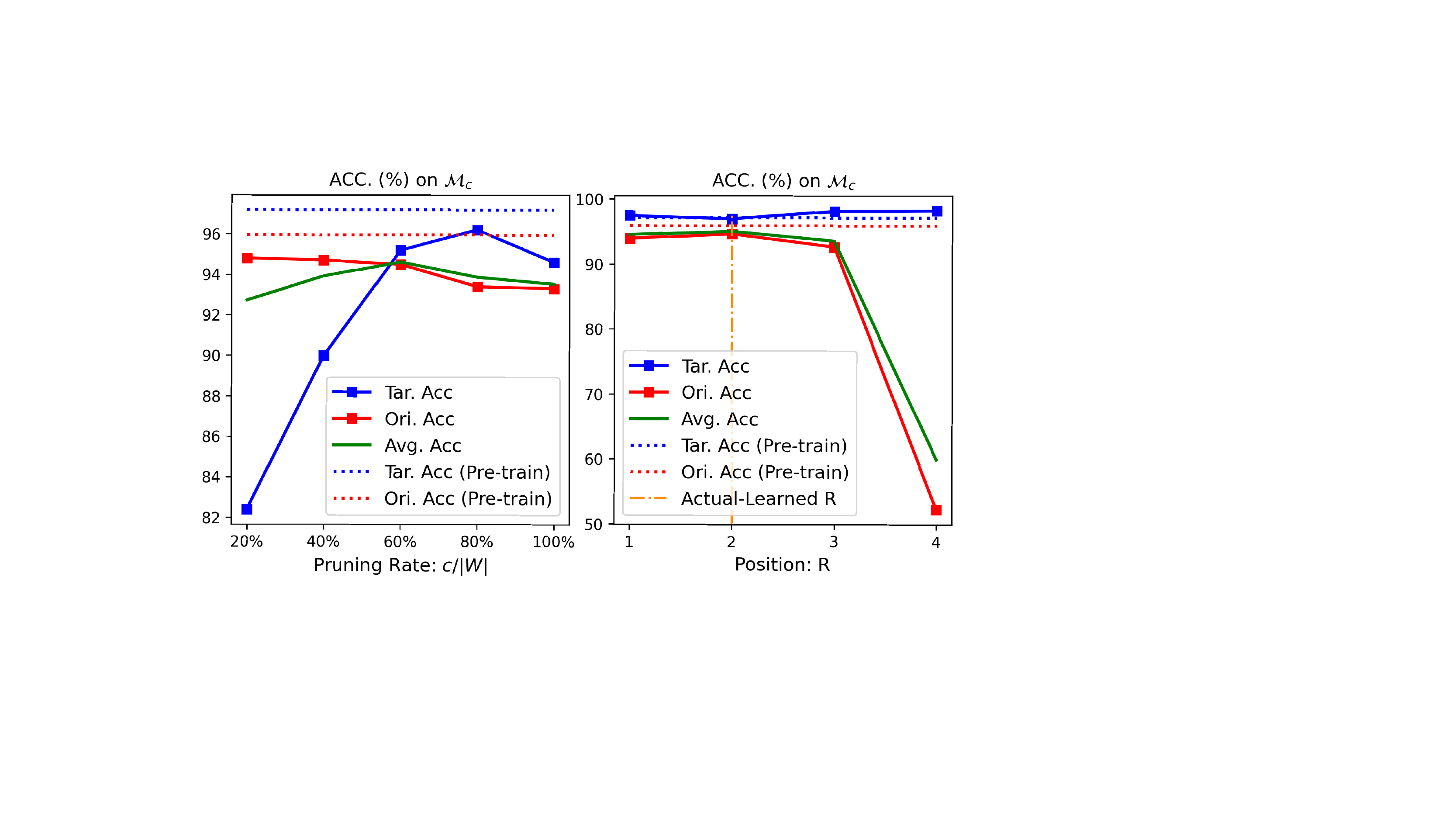}
\caption{The performance on partial network cloning with different scales of the transferable module.}
\label{fig:size}
\end{figure}

\subsubsection{More Analysis of the Transferable Module}
\textbf{How does the scale of the transferable module influence the cloning performance?}
The transferable module can be denoted as $\mathcal{M}_f\leftarrow M \cdot W_s^{[R:L]}$.
And the scale of the transferable module is influenced by two factors, which are the selection function $M$ and the position parameters $R$. 
We explore the influence of the scale on the CIFAR-10, with the same setting from Table~\ref{tab:overall} of cloning small functionality.
The selection function $M$ is directly controlled by the masking rate ${c}/{|W|}$ ($0\le l<L$, defined in Eq.~\ref{eq:locloss}), where larger $c^l$ makes larger transferable modules, shown in Fig.~\ref{fig:size} (left).
As can be observed from the figure, the accuracy of the original functionality (`Ori. Acc') slightly decreases with larger $\mathcal{M}_f$. While larger $\mathcal{M}_f$ doesn't ensure higher accuracy of the to-be-cloned function (`Tar. Acc', first increase and then drop), indicating that the appropriate localization strategy on the source instead of inserting the whole source network benefits a lot.

The position parameter $R$ ($0\le l<L$) is learned in the insertion process, here we show the performance for $R = 1\sim 4$, which further shows the validation of our selection strategy. Bigger $R$ makes smaller transferable modules, the accuracy based on which is shown in Fig.~\ref{fig:size} (right). The accuracy on the to-be-cloned set (`Tar. Acc') doesn't largely influenced by it, while it does influence the accuracy on the original set (`Ori. Acc') a lot.
Notably, $R=2$ is the position learned in the insertion process, which shows to the best according to the average accuracy (`Avg. Acc').  

\begin{figure}[t]
\centering
\includegraphics[scale = 0.5]{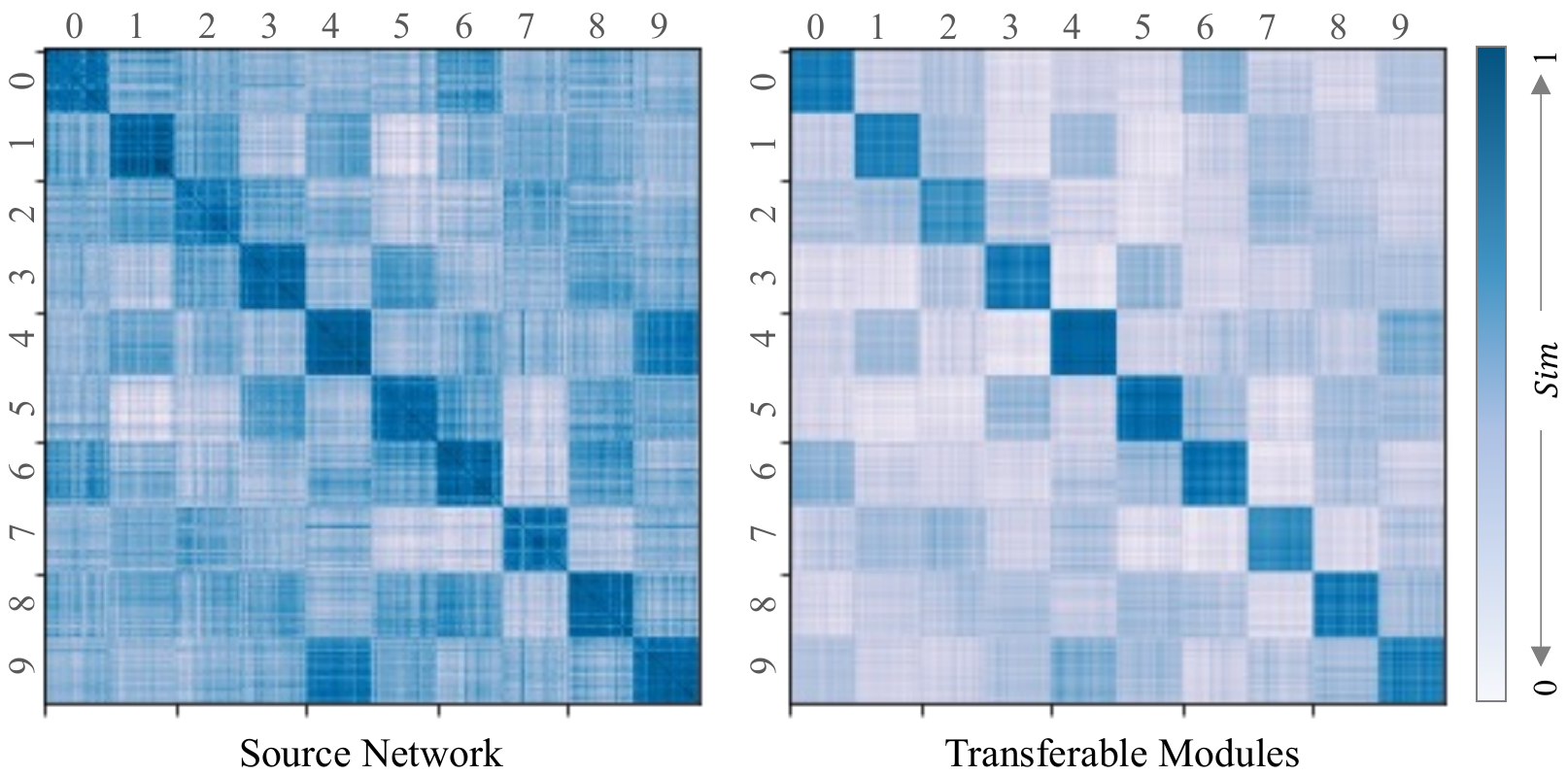}
\caption{The similarity matrix maps computed on the source network and on the transferable modules on MNIST dataset. Deeper color indicates higher similarity.}
\label{fig:sim}
\end{figure}

\textbf{Has the transferable module been explicitly localized?} 
For evaluating the quality of the transferable module on whether the learned localization strategy $Local(\cdot)$ has successfully selected the explicit part for the to-be-cloned functionality or not, we compute the similarity matrix for the source network and the transferable module, which is displayed in Fig.~\ref{fig:sim}.
The comparison is conducted on the MNIST dataset, which is split into 10 sub-datasets ($D_0\sim D_9$, according to the label) and each time we localize one-label functionality from the source, thus obtaining 10 transferable modules ($\mathcal{M}_f^0\sim \mathcal{M}_f^9$). 
For the source network, we compute $Sim(\mathcal{M}_s|D_i,\mathcal{M}_s|D_j)$ for each sub-dataset pair. It could be observed from Fig.~\ref{fig:sim} (left) that each functionality learned from each sub-dataset are more or less related, showing the importance of the localization process to extract local functionality from the rest.
For the transferable module, we compute $Sim(\mathcal{M}_s|D_j,\mathcal{M}^i_f|D_i)$ . Fig.~\ref{fig:sim} (right) shows that the transferable module has high similarity with the source network on the to-be-cloned sub-dataset, and its relation with the rest sub-dataset is weakened (Non-diagonal regions are marked in lighter color than the matrix map of the source network).
Thus, the conclusion can be drawn that \textit{the transferable module successfully models the locality performance on the to-be-cloned task set}, proving the correctness of the learned $Local(\cdot)$.

\begin{table}[t]
\centering
\caption{The cloning performance evaluation on heterogeneous model pairs on CIFAR-10 and CIFAR-100 datasets.}
\scriptsize
\begin{tabular}{p{14mm}<{\centering}p{9mm}<{\centering}p{9mm}<{\centering}ccc}
\toprule
\textbf{Source}& & &  \multicolumn{3}{c}{\textbf{Acc. on $\mathcal{M}_c$}} \\ \cmidrule(r){4-6}
\textbf{(Target)}&\textbf{Dataset}& \textbf{Method} & \textbf{Ori. Acc} & \textbf{Tar. Acc}& \textbf{Avg. Acc} \\\midrule \midrule
ResNet-18&CIFAR10& Pre-train & 88.1 &95.9 &- \\ 
/&CIFAR10& Continual &75.3 & 92.8 & 78.2 \\
(CNN)&CIFAR10& PNC & 86.9 & 90.3& \textbf{87.5}\\ \midrule
ResNet-18&CIFAR10&Pre-train& 92.6 &95.9&-  \\ 
/&CIFAR10& Continual& 85.0 &96.7&87.0\\
(ResNetC-20)&CIFAR10& PNC & 92.1  & 93.5&\textbf{92.3}\\ \midrule
ResNet-18 &CIFAR100&Pre-train &72.9& 78.1&-\\
/&CIFAR100& Continual& 66.7&79.9&68.9\\
(MobileNetV2)&CIFAR100&PNC&70.8 &76.1& \textbf{71.7}\\\midrule 
ResNet-18&CIFAR100&Pre-train &74.8& 78.1&-\\
/&CIFAR100& Continual & 63.8& 82.3& 66.9\\
(ShuffleNetV2)&CIFAR100&PNC&72.9&77.1& \textbf{73.6} \\
\bottomrule
\end{tabular}
\vspace{-1.5em}
\label{tab:Heterogeneous Models}
\end{table}

\subsubsection{Cloning between Heterogeneous Models}
Here we evaluate the effectiveness of the proposed PNC between heterogeneous models.
The experiments are conducted on the CIFAR-10 and CIFAR-100 datasets, where 20\% of the functionalities are cloned from the source network to the target network. The results are depicted in Table~\ref{tab:Heterogeneous Models}. In the figure, we compare our PNC with the original pre-trained models and the network trained in continual learning setting.
As can be seen in the figure, cloning shows superiority between similar architectures of the source and target pair (`ResNet-18 (ResNetC-20)' has higher accuracies than `ResNet-18 (CNN)').

\subsubsection{Comparing with Incremental Learning}
The proposed partial network cloning framework can be also conducted to tackle incremental learning task.
The comparisons are made on the CIFAR-100 dataset when using ResNet-18 as the base network. We pre-train the target network with the first 50 classes and continually add the rest from the source network with different class-incremental step $s$.
The comparative results with some classic incremental learning methods are displayed in Table~\ref{tab:il}, where we compare PNC with the regularization- rehearsal- and the architecture- based continual learning methods, and show its superior in classification performance. More insight analysis and comparison with incremental learning are included in the supplementary.

\begin{table}[t]
\centering
\scriptsize
\caption{Comparative experimental results on incremental learning CIFAR-100 dataset.}
\label{tab:recoverorder}
\begin{tabular}{lccccc}
\toprule
\textbf{Method}& \textbf{Description} &$s=5$& $s=10$ &$s=20$&$s=50$\\ \hline\hline
LwF~\cite{Li2016LearningWF} & Regularization&29.5 &  40.4 & 47.6& 52.9  \\
iCaRL~\cite{Rebuffi2017iCaRLIC}& Rehearsal &57.8 & 60.5  & 62.0  & 61.8  \\
EEIL~\cite{Castro2018EndtoEndIL}& Rehearsal &63.4  & 63.6  & 63.7  & 60.8 \\
BiC~\cite{Wu2019LargeSI}& Architecture &60.1 & 60.4&68.9 & 70.2\\
PNC(ours)& Architecture &\textbf{71.5}&\textbf{73.6}&\textbf{75.2}&\textbf{74.2} \\
\bottomrule
\end{tabular}
\vspace{-1.5em}
\label{tab:il}
\end{table}

\section{Conclusion}
In this work, we study a new knowledge-transfer task,
termed as \emph{Partial Network Cloning}~(PNC),
which clones a module of parameters
from the source network
and inserts it to the target
in a copy-and-paste manner.
Unlike prior knowledge-transfer settings
the rely on updating parameters 
of the target network,
our approach preserves the parameters
extracted from the source
and those of the target
unchanged.
Towards solving PNC, 
we introduce an effective learning
scheme that jointly conducts
localizing  
and  insertion,
where the two steps reinforce each other.
We show on several datasets that our method yields encouraging results on both the accuracy and locality metrics,
which consistently outperform the results
from other settings.

\section*{Acknowledgements}
This work is supported by the Advanced Research and Technology Innovation Centre (ARTIC), the National University of Singapore (project number:~A-0005947-21-00, project reference:~ECT-RP2),
and National Research Foundation, Singapore under its Medium Sized Center for Advanced Robotics Technology Innovation.

\end{document}